\documentclass[10pt,twocolumn,letterpaper]{article}

\usepackage{wacv}
\usepackage{times}
\usepackage{epsfig}
\usepackage{graphicx}
\usepackage{amsmath}
\usepackage{amssymb}
\usepackage{makecell}
\usepackage[dvipsnames]{xcolor}
\usepackage[outdir=./images]{epstopdf}

\usepackage[pagebackref=true,breaklinks=true,letterpaper=true,colorlinks,bookmarks=false]{hyperref}

\wacvfinalcopy %

\def\wacvPaperID{544} %

\ifwacvfinal\fi
\begin{document}

\title{Exploring 3 R's of Long-term Tracking: Re-detection, Recovery and Reliability}    

\author{Shyamgopal Karthik\quad
Abhinav Moudgil\quad
Vineet Gandhi\\
Center for Visual Information Technology, Kohli Center on Intelligent Systems\\
IIIT-Hyderabad, India\\
{\tt\small \{shyamgopal.karthik@research., abhinav.moudgil@research., vgandhi@\}iiit.ac.in}}

\maketitle
\ifwacvfinal\thispagestyle{empty}\fi

\begin{abstract}
Recent works have proposed several long term tracking benchmarks and highlight the importance of moving towards long-duration tracking to bridge the gap with application requirements. The current evaluation methodologies, however, do not focus on several aspects that are crucial in a long term perspective like Re-detection, Recovery, and Reliability.  In this paper, we propose novel evaluation strategies for a more in-depth analysis of trackers from a long-term perspective. More specifically, (a) we test re-detection capability of the trackers in the wild by simulating virtual cuts, (b) we investigate the role of chance in the recovery of tracker after failure and (c) we propose a novel metric allowing visual inference on the ability of a tracker to track contiguously (without any failure) at a given accuracy. We present several original insights derived from an extensive set of quantitative and qualitative experiments. 

\end{abstract}
\section{Introduction}

Visual tracking is a fundamental problem in computer vision and has rapidly progressed in the recent past with the onset of deep learning. However, progress is still far from matching practitioner needs, which demands consistent and reliable long-duration tracking. Interestingly, most existing works evaluate their performance on datasets consisting of multiple short clips. For instance, the most commonly used OTB dataset has an average length of about 20 seconds~\cite{WuLimYang13} per clip. Work by Moudgil and Gandhi~\cite{moudgil2017long} observed a sharp performance drop when the trackers were evaluated on long sequences. Following works~\cite{fan2019lasot, valmadre2018long, lukevzivc2018now} also make similar observations and suggest that we need alternate ways to evaluate and analyze long term tracking performance. 

\begin{figure}[t]
\begin{center}
   \includegraphics[width=0.8\linewidth]{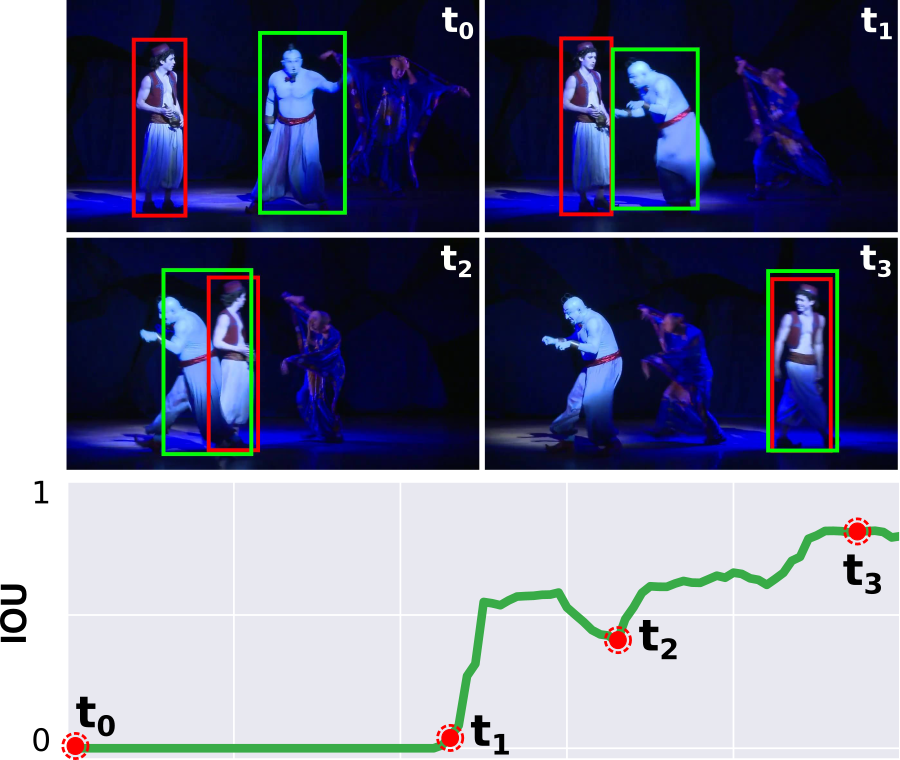}
\end{center}
   \caption{A typical example of a chance based recovery in Alladin sequence from TLP~\cite{moudgil2017long} dataset. \textcolor{YellowGreen}{SiamRPN} (green) is tracking the incorrect object and has zero overlap with the \textcolor{red}{target} (red) in the start. It switches to tracking the target when they pass through each other. We study such chance based recoveries in long-term setting both qualitatively and quantitatively. Best viewed in colour.}
\label{fig:teaser}
\end{figure}

These works~\cite{moudgil2017long, fan2019lasot, valmadre2018long, lukevzivc2018now} indicate that three properties are crucial for an improved long term tracking performance. First is the ability to re-detect the target if it is lost. \emph{Re-detection} is crucial to handle situations where the target object goes out of the frame and reappears. It is also essential to re-initiate tracking when the target object is lost due to occlusions or momentary tracking failures. The second key aspect is the ability of the tracker to distinguish between the actual target and distractor or background clutter. This aspect is vital for consistency in tracking as well as for \emph{recovery} from failures. Figure \ref{fig:teaser} illustrates an example where chance plays a crucial role in recovery. We believe that scrutinizing the nature of failures and recoveries will aid improved tracking performance. The third key aspect is \emph{Reliability}, which connects to the ability for consistent and contiguous tracking. Contiguity suggests the ability of the tracker to track for a long duration without any failure. Consistency indicates the accuracy of tracking over time. Tracking in the long-duration video allows us to study factors like a slow accumulation of error which is difficult to observe in short sequences. Several applications like video surveillance or virtual camera simulation from static camera~\cite{gandhi2014multi} require precise tracking for long time. Surprisingly, none of the current evaluation strategies focus on these three crucial aspects of \emph{Re-detection}, \emph{Recovery}, and \emph{Reliability}.

For instance, the most prevalent metrics are Success and Precision plots, which measure the number of frames with Intersection Over Union (IoU) greater than a threshold and the mean distance from the center of the ground truth respectively. Both these metrics do not reflect anything specific about the 3R's. Recent work by Lukezic~\etal~\cite{lukevzivc2018now} studied the efficacy of the search region expansion strategy of different trackers. However, the evaluation is performed in a synthetic experimental setup (designed by padding with gray values) and may not be an indicator of performance in real-world scenarios. Valmadre~\etal~\cite{valmadre2018long} improves the evaluation strategy by explicitly handling the cases where the target is not visible/absent from the frame. Other recent efforts~\cite{moudgil2017long, fan2019lasot} identify the aforementioned key issues, but they do not provide any way to evaluate these properties comprehensively. 

In this work, we propose two novel evaluation metrics focused on the re-detection ability and the aspect of continuous and consistent long term tracking. Furthermore, we present more in-depth insights into the failure and recovery of different trackers, explicitly addressing the role of distractors. Since shorter sequences are inappropriate to address these concerns, we use the Track Long and Prosper (TLP)~\cite{moudgil2017long} dataset for the experiments. The main advantage of TLP is that the average sequence length is longest among other densely annotated datasets~\cite{huang2018got,fan2019lasot,lukevzivc2018now}. Long duration videos present cases of multiple failures and recoveries for each video, which allows for a deeper analysis. Our contributions include:

\begin{enumerate}
    \item We propose a novel way to quantitatively evaluate the re-detection abilities of a tracker by simulating cuts (an abrupt transition from a frame to another) in original videos. We propose a method to search challenging locations for placing the cut by minimizing the Generalized IoU~\cite{Rezatofighi_2019_CVPR} between the ground truth bounding boxes in the frame before and after the cut. Different trackers are then evaluated on their ability to recover/re-detect, and the time they take to recover. 
    \item We formally study the chance factor in recoveries post-failure. We analyze the role of distractors in failures and recoveries and the co-incidences which aid tracking. For example, it often happens in long sequences that tracker loses the target at some location and freezes there. If by chance the target passes the same location (after a while), the tracker starts tracking it again. Our study aims to quantify such behavior.
    \item We propose 3D Longest Subsequence Measure (3D-LSM), as a novel metric for quantifying tracking performance. It measures the longest contiguous sequence successfully tracked at a given precision and allowed failure tolerance. The 3D-LSM allows for a direct visual interpretation of tracking results in the form of a 2D image.
\end{enumerate}

\section{Related Work}
\begin{figure*}
\centering
\begin{minipage}{.48\textwidth}
\begin{center}
   \includegraphics[width=\linewidth]{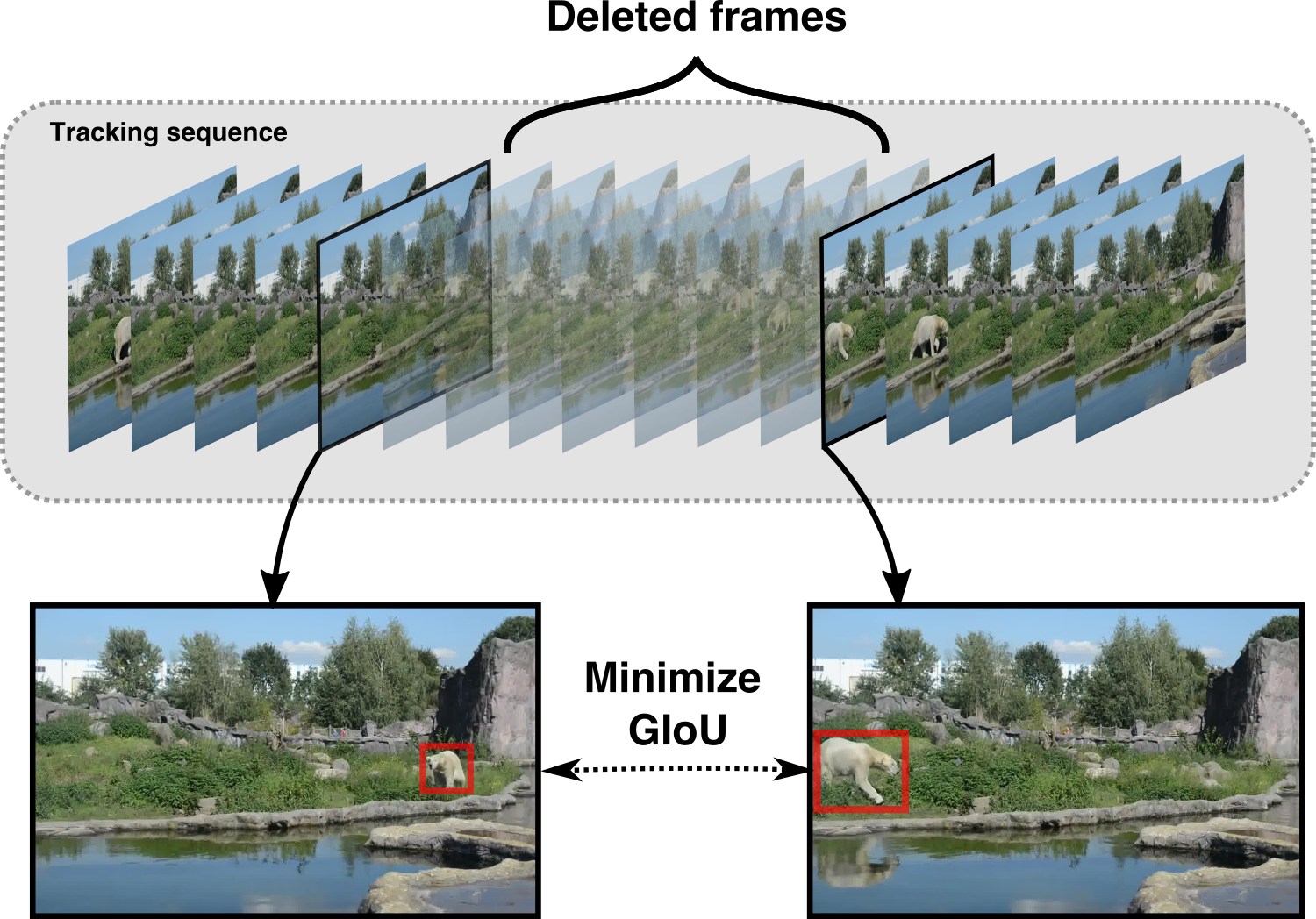}
\end{center}
\end{minipage}%
\begin{minipage}{.32\textwidth}
\begin{center}
   \includegraphics[width=0.83\linewidth]{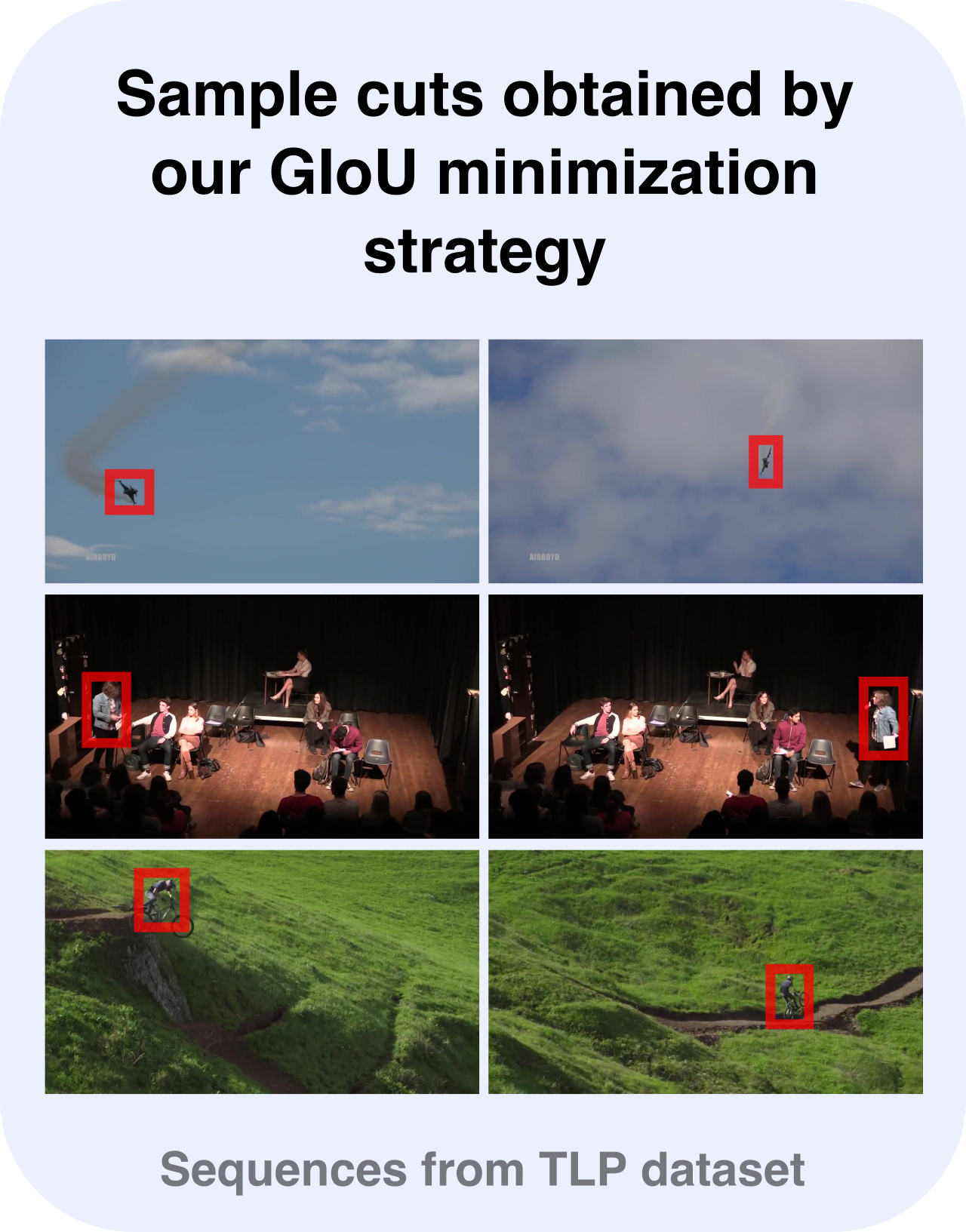}
\end{center}
\end{minipage}%
\vspace{0.5em}
\caption{A cut is introduced by removing a set of contiguous frames from a tracking sequence. This introduces a sudden change of position of the ground truth object as shown in the left diagram. The red bounding box shows the position of the target object, before and after the cut. We maximize the amount of target shift by minimizing the GIoU \cite{Rezatofighi_2019_CVPR} between these bounding boxes. We evaluate the trackers ability to re-detect the object after the cut. Few more examples from TLP dataset with simulated cuts are shown on the right.}
\label{fig:cut}
\end{figure*}
\textbf{Tracking Datasets:} There are a large variety of datasets for object tracking. Most commonly used datasets are OTB50\cite{WuLimYang13} and OTB100\cite{Wu2015}. They consist of short videos from generic real-world scenarios. ALOV300\cite{smeulders2013visual} increased diversity by including 300 short sequences. The average video length in ALOV dataset is only 9 seconds. NFS\cite{galoogahi2017need} dataset included sequences recorded at high frame rate (240fps). UAV\cite{mueller2016benchmark} introduced a dataset from sequences shot from an aerial vehicle. Moudgil and Gandhi\cite{moudgil2017long} TLP dataset with 50 sequences, focusing on long-duration tracking (significantly increasing length of individual sequences). LTB35\cite{lukevzivc2018now} and OxUvA\cite{valmadre2018long} then followed emphasizing the need to focus on long term tracking. LaSOT\cite{fan2019lasot} significantly increased the size of the dataset with over 1000 sequences. GOT10k\cite{huang2018got} then followed by proposing a dataset with 10000 sequences, including objects from 563 different classes.

\textbf{Tracking Methods:} We list some notable attempts which led to significant progress in long term tracking. Collins~\etal~\cite{collins2005online} proposed the idea of using the neighborhood around the ground truth for discriminative feature learning. This idea was later formalized into tracking by detection frameworks~\cite{babenko2009visual}. Kalal~\etal~\cite{kalal2011tracking} proposed TLD framework of learning detector from initial tracking, maintaining the confidence of local tracking based on feature point tracks, and switching to detection in low confidence scenarios. TLD tracker was one of the first attempts to elegantly handle the re-detection problem, which is crucial for long term tracking. The consistency aspect was then improved by employing an ensemble of classifiers~\cite{zhang2014meem}. These methods maintain several weak classifiers, often initiated at different checkpoints to account for appearance variations of the target.

Another popular direction is Discriminative Correlation Filter (DCF) based tracking~\cite{bolme2010visual, danelljan2015learning}. These methods exploit the properties of circular correlation (efficiently performed in Fourier domain) for training a regressor in a sliding-window fashion. Recent progress in DCF's is driven by integrating multi-resolution shallow and deep features maps to learn the correlation filters~\cite{danelljan2017eco,bhat2018unveiling,valmadre2017end}. Another fundamental contribution is the use of Siamese networks for visual object tracking~\cite{bertinetto2016fully,held2016learning}. The GOTURN tracker~\cite{held2016learning} uses the Siamese architecture to directly regress the bounding box locations given two cropped images from previous and current frames. The SiamFC tracker~\cite{bertinetto2016fully} transforms the exemplar image and the large search image using the same function and outputs a map by computing similarity in the transformed domain. These efforts~\cite{bertinetto2016fully,held2016learning} do not include any online updates and are extremely efficient in terms of computation. The Siamese framework was further augmented by employing Region Proposal Networks (RPN)\cite{li2018siamrpn++,Li_2018_CVPR} which significantly improves the accurate prediction of scale and aspect ratio of the bounding boxes.

Another pioneering effort came from Nam~\etal~\cite{nam2016learning}, who introduced the idea of treating the tracking problem as classifying candidate windows sampled around the previous target region. Several recent efforts have explored the variations of the Tracking Learning Detection (TLD) framework. Nebehay \etal~\cite{nebehay2015clustering} proposed a mechanism to drift by filtering outlier correspondences. A combination of short term CF tracker with additional components (e.g., an explicit re-detection module) have been explored~\cite{lukevzivc2018fucolot,ma2015long}. Zhang \etal~\cite{zhang2018learning} employed an offline trained regression network as the short-term component and an online-trained verification network to detect tracking failure and start image wide detection. Yan~\etal~\cite{iccv19_SPLT} show significant computational improvements by replacing the online verification network with an offline trained Siamese verification network.

\textbf{Tracking Metrics:} Early works relied on the precision metric\cite{babenko2009visual,WuLimYang13} for quantifying the tracking performance, which computes the pixel distance between the center of the ground truth and the prediction. This was convenient since it required only annotating the center of the target and not the whole bounding box. However, since this does not account for the scale and aspect ratio, the success metric\cite{WuLimYang13} was introduced. It measures the percentage of frames where the Intersection Over Union (IOU) of the predicted and ground truth bounding boxes is more than a threshold. Failure rate~\cite{kristan2010two} was then introduced to address the continuity and consistency aspect of tracking. In failure rate measure, a manual operator reinitializes the tracker upon every failure. The number of required manual interventions per frame is recorded as the quantitative measure. However, due to the need of manual interventions, it is unscalable for long sequences (in large datasets). For a more detailed review and analysis of metrics for short-term tracking, we would refer the reader to work by Cehovin~\etal~\cite{vcehovin2014my}.

A few evaluation metrics have been proposed targeting long-duration tracking. Valmadre \etal\cite{valmadre2018long} introduced True Positive Rate(TPR), True Negative Rate(TNR) and took their geometric mean. To have a single representative metric accounting for the trackers which do not predict absent labels, they proposed a modified metric called maximum geometric mean metric. However, the metric is biased towards the ability of a tracker to predict absent labels.

Lukezic \etal \cite{lukevzivc2018now} introduced tracking recall and precision and used this to give a tracking F1 score. However, their definition of a long term tracker is limited to the ability of a tracker to predict absence, and the proposed metric does not focus on the continuity and consistency aspect of tracking. We believe the ability to track for long-duration consistently even when the target object is always present has been overlooked in these previous efforts~\cite{lukevzivc2018now, valmadre2018long}. Lukezic \etal also proposed an experiment to quantify the re-detection ability of a tracker. However, their experiment mainly focuses on the search strategy with no appearance changes. Here, we seek to quantify the re-detection ability in the wild. Moudgil and Gandhi~\cite{moudgil2017long} proposed the Longest Subsequence Measure (LSM), which quantifies the longest contiguous segment successfully tracked in the sequence. Here, we propose an extension of it called 3D-LSM, which allows comparing trackers visually.

\section{Re-detection in the Wild}
This experiment is designed to quantify a tracker's ability to re-detect the object after it is lost (either because the target goes of the view or due to momentary failures).

\textbf{Setup:} We select a segment from a sequence, and delete it, thereby introducing a cut (illustrated in Figure~\ref{fig:cut}). We evaluate the tracker's performance on the segment after the cut to evaluate the re-detection ability of the tracker. For each sequence from the TLP dataset, we cut a segment that minimizes the Generalized IoU (GIoU)~\cite{Rezatofighi_2019_CVPR} between the target bounding boxes before and after the cut. GIoU allows capturing the ``distance" between bounding boxes in a generic way which implicitly takes into account various factors like center distance, scale, and aspect ratio. The duration of the cut is fixed to 300 frames. We empirically find that 300 frames allow the target to move far away from the tracker's search region without significantly varying the other aspects in the scene. Keeping a similar context around the target helps to keep the focus on the re-detection ability (the context can change dramatically in long sequences if the length of the omitted sequence is large).  The proposed re-detection scheme is quite general and can be applied even on datasets that do not have target disappearances at all.

\textbf{Evaluation:} In all the experiments the tracker is initialized 100 frames before the cut. We choose 100 frames so that the tracker starts stable tracking before the cut. It also allows trackers with online updates to build a reasonable representation of the target object. We also make sure that there are no critical challenges in this duration of 100 frames such as heavy occlusion, clutter, etc. to avoid tracker failure in these 100 frames. After the cut, the tracker is continued to run on the sequence for another 200 frames and its performance on this segment is evaluated. We define ``recovery" when the IoU of the tracker with the target reaches 0.5. To make a relative comparison of the trackers on the re-detection task, we report the following metrics. 
\begin{enumerate}
    \item Total number of sequences (out of the total 50 TLP sequences) in which a tracker is able to recover within the remaining 200 frames.
    \item Total number of sequences where the recovery is ``quick," i.e., the recovery happens within 30 frames (1 second).
    \item Average number of frames a tracker takes to recover successfully.
\end{enumerate}

We perform this experiment on TLP dataset with the following trackers: SPLT~\cite{iccv19_SPLT}, MBMD~\cite{zhang2018learning}, FuCoLoT~\cite{lukevzivc2018fucolot}, ATOM~\cite{Danelljan_2019_CVPR}, MDNet~\cite{nam2016learning}, SiamRPN~\cite{Li_2018_CVPR}, ECO~\cite{danelljan2017eco}, CMT~\cite{nebehay2015clustering}, LCT~\cite{ma2015long}, and TLD~\cite{kalal2011tracking}. SPLT, MBMD, FuCoLoT, CMT, LCT and TLD are long-term trackers with explicit re-detection ability; ATOM is the current top performing tracker on the long-term benchmark LaSOT, while MDNet, SiamRPN and ECO are the top performing trackers on other benchmarks~\cite{moudgil2017long, Wu2015,kristan2018sixth}. This selection presents all the prevalent tracking approaches: correlation filter based trackers~\cite{danelljan2017eco, ma2015long,lukevzivc2018fucolot}, end to end classification with online updates~\cite{nam2016learning}, offline trained Siamese trackers with region proposals~\cite{Li_2018_CVPR}, low level feature tracking with online learned detector~\cite{kalal2011tracking,nebehay2015clustering} and combination of multiple offline/online trained components~\cite{Danelljan_2019_CVPR,zhang2018learning,iccv19_SPLT}. The same set of trackers are used in all the following experiments as well.

\begin{table}
\footnotesize
\begin{center}
\begin{tabular}{|c|c|c|c|}
\hline
\thead{Tracker} & \thead{Quick \\recoveries $\uparrow$ } & \thead{Total \\recoveries $\uparrow$} & \thead{Avg. recovery \\ length (\# frames) $\downarrow$} \\
\hline
SPLT~\cite{iccv19_SPLT} & \textbf{20} & \textbf{36} & 19\\
\hline
FuCoLoT~\cite{lukevzivc2018fucolot} & 10 & 33 & 55\\ 
\hline
MBMD~\cite{zhang2018learning} & 15 & 27 & 28 \\
\hline
ATOM~\cite{Danelljan_2019_CVPR} & 12 & 25 & 34 \\
\hline
CMT~\cite{nebehay2015clustering} & 9 & 14 & 22\\
\hline
TLD~\cite{kalal2011tracking} & 6 & 10 & \textbf{8}\\
\hline
MDNet~\cite{nam2016learning} & 5 & 13 & 48\\
\hline
ECO~\cite{danelljan2017eco} &  4 & 7 & 28 \\
\hline
SiamRPN~\cite{Li_2018_CVPR} & 2 & 7 & 39 \\
\hline
LCT~\cite{ma2015long} & 2 &  7 & 143\\
\hline
\end{tabular}
\end{center}
\caption{Results for the re-detection experiment (out of 50 sequences).}
\label{table:table1}
\end{table}

\begin{figure}[t]
\begin{center}
\includegraphics[width=\linewidth]{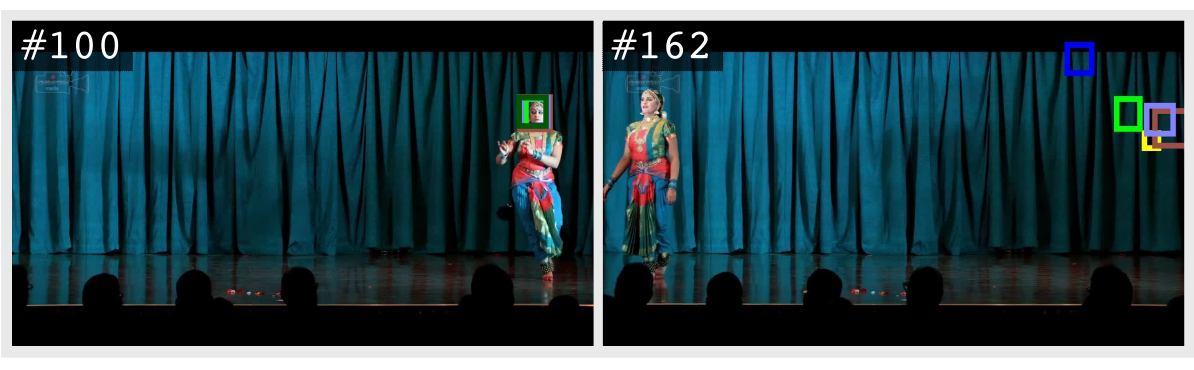}
\end{center}
  \caption{The figure illustrates a simulated cut in the Bharatanatyam sequence from TLP dataset. The cut can be seen as a representation of a situation where the performer exits the stage and enters from another end. None of the 6 trackers was able to recover in this sequence, even with the exact same background and a single target object.}
\label{fig:bharat}
\end{figure}

\textbf{Results and Discussion:} Our results are summarized in Table \ref{table:table1}. SPLT gives the best results, followed by FuCoLoT and MBMD. Since the base framework of SPLT and MBMD is the same as SiamRPN, the significant improvements (from SiamRPN to SPLT) can be attributed to the additional verification and re-detection module. An explicit re-detection module also improves CF-based trackers (as seen in FuCoLoT). CMT and TLD dominate in re-detection experiments studied in previous works~\cite{lukevzivc2018now}; however, they give poor results in our experiment. We empirically observe that CMT and TLD fail to adapt to appearance changes that occur before and after the cut, possibly because of the weak appearance model used in the detector. Adapting to appearance changes during re-detection is essential in real-world settings and previous synthetically designed experiments ~\cite{lukevzivc2018now} do not account for this aspect. Other trackers like ECO, MDNet, and SiamRPN are limited by their search range and only recover if the target object comes within their search range after the cut. ATOM, on the other hand, uses a larger search area (25 times the area of target object bounding box) and hence recovers more often. Qualitatively, we observe sequences with background clutter or with distractors prove to be the most challenging for all the trackers. Re-detection results are also poor on targets that are small in size.

\section{Recovery by Chance}

\begin{figure}[t]
\begin{center}
   \includegraphics[width=0.8\linewidth]{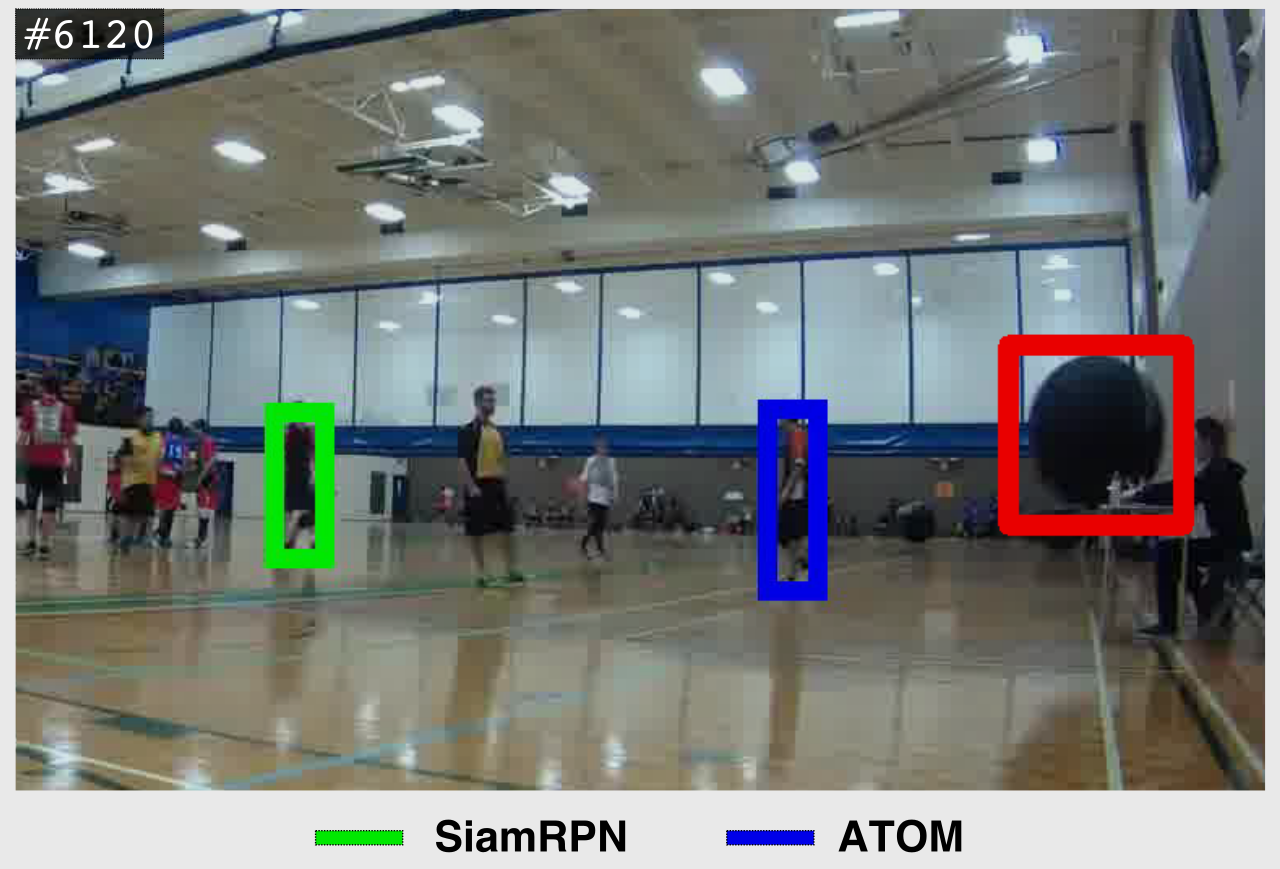}
\end{center}
   \caption{An example from TLP~\cite{moudgil2017long} Kinball1 sequence where the tracking \textcolor{red}{target} is black ball. Both \textcolor{YellowGreen}{SiamRPN} and \textcolor{blue}{ATOM} end up tracking objects of totally different class i.e. human which is also significantly different in appearance from the given target.}
\label{fig:distractor}
\end{figure}

In this section, we investigate the role of chance in tracker recovery post-failure. Interestingly, most of the evaluation metrics~\cite{lukevzivc2018now,kristan2018sixth,Wu2015} do not take this into account, and we believe that to design better long-term trackers, it is essential to scrutinize the nature of recovery. More specifically, we analyze two scenarios that frequently occur in long sequences (a) the tracker starts tracking an alternate object and recovers back when it interacts with the target and (b) tracker freezes somewhere in the background and resumes tracking when the target passes through it.

\subsection{Recovery by Tracking Alternate Object}
We first investigate the cases when the recovery occurs while tracking an alternate object (distractor). We consider distractors of both the same class as well as other classes. The recovery here occurs only because of the interactions between the objects in the scene. An example of this kind of recovery is illustrated in Fig \ref{fig:teaser}. 

However, directly evaluating the role of distractors is challenging because single object tracking benchmarks~\cite{Wu2015, moudgil2017long, kristan2018sixth} do not have annotations for multiple objects. We exploit the effectiveness of modern object detectors to resolve this concern.  While an object detector would not be accurate enough to be treated as ground truth for bounding boxes for alternate objects, it would still allow us to draw useful insights. Moreover, the results may vary when a different object detector is being used. Hence, the evaluation presented in this section is not intended to serve as a metric. Nonetheless, it presents important insights into the role of distractors in tracking performance, which is further highlighted by qualitative results presented in the supplementary material.

We select 16 out of the 50 sequences from the TLP dataset where distractors are present, and the target interacts with them. We run YOLOv3~\cite{redmon2018yolov3,Redmon_2016_CVPR} on these sequences to obtain all object annotations. We compute and study the following aspects:
\begin{itemize}
    \item The percentage of frames where the tracker is tracking (IoU $\ge$ 0.5) an alternate object and has zero overlap with the target (averaged over the selected 16 sequences).
    \item The recoveries that occur while the tracker is tracking an alternate object (IoU with alternate object $\ge$ 0.5).  We define recovery if the IoU with the ground truth becomes nonzero and maintains a non zero value for the next 60 frames. We present the number of recoveries per sequence for each tracker.
    \item The performance drop that occurs if we zero out the performance after the first instance of such a recovery.
\end{itemize}

\begin{table*}
\footnotesize
\begin{center}
\begin{tabular}{|c|c|c|c|c|}
\hline
\thead{Tracker} & \thead{Mean \% of frames where \\alternate object was \\ tracked } & \thead{Avg. no. of \\ Recoveries} & \thead{Original \\ Performance \\ (Success Metric)} & \thead{Reduced\\ Performance \\ (Success Metric)} \\
\hline 
SPLT~\cite{iccv19_SPLT} & 20.92\% & 8.93 & 36.19 & 17.75 \\
\hline
MBMD~\cite{zhang2018learning} & 19.03\% & 5.75 & 32.99 & 15.32\\
\hline
ATOM~\cite{Danelljan_2019_CVPR} & 13.73\% & 5.68 & 31.42 & 18.81 \\
\hline
SiamRPN~\cite{Li_2018_CVPR} & 14.92\% & 5.18 & 43.80 & 25.19 \\
\hline
FuCoLoT~\cite{lukevzivc2018fucolot} & 1.8\% & 1.18 & 23.33 & 19.34\\
\hline
MDNet~\cite{nam2016learning} & 1.57\% & 0.68 &  40.77 & 38.45\\
\hline
CMT~\cite{nebehay2015clustering}  & 2.77\% & 0.68 & 8.76 & 7.98 \\
\hline
ECO~\cite{danelljan2017eco} &  3.8\% & 1.5 &  22.38 & 19.68\\
\hline
LCT~\cite{ma2015long} & 1.38\% &  0.81 &  11.21 & 10.42\\
\hline
TLD~\cite{kalal2011tracking} & 0.58\% & 0.12 & 7.02 & 4.32\\

\hline
\end{tabular}
\end{center}
\caption{Results for the analysis of distractor enabled recoveries.}
\label{table:table2}
\end{table*}

\textbf{Results and Discussion:}  The results are shown in Table~\ref{table:table2}. SPLT, MBMD, ATOM, and SiamRPN track an incorrect object for more than 13\% of the frames on average in a sequence, which is an exceedingly high number. The behavior possibly stems from the nature of their design which looks for ``objectness" i.e., the potential bounding boxes in the neighborhood. SPLT and ATOM despite employing hard negative mining strategies while training are prone to tracking alternate objects.  Most trackers are highly susceptible to intraclass variations like the color, pose, clothing, etc. and keep on confusing on cases like two boxers in a ring or two nearby cars on a highway. The confusion among different classes is also observed (Fig~\ref{fig:distractor}). Interestingly, the trackers which perform online model updates  (MDNet, FuColoT, ECO, CMT and TLD) are less susceptible to track an alternate object.

In the last two columns of Table~\ref{table:table2} we present the success metric of the listed trackers on the selected 16 sequences and the reduced performance computed by setting the IOU scores to zero after the first chance-based recovery. The reduced performance is indicative of the worst-case performance, i.e., if a chance-based recovery never happened. We observe a significant drop in the case of SPLT, MBMD, ATOM, and SiamRPN. The performance drop for other trackers is also significant in the context of their overall tracking performance (for example, TLD's performance drops by more than 35\%).

\subsection{Recovery With No Motion}
\begin{figure}[t]
\begin{center}
   \includegraphics[width=0.8\linewidth]{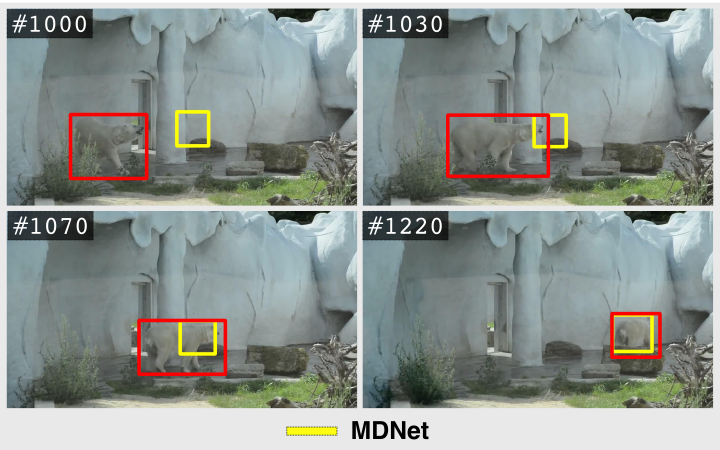}
\end{center}   \caption{An example of a recovery where the tracker does not move at all, but the ground truth (red) falls right into the tracker's (yellow) prediction }
\label{fig:static}
\label{fig:static1}
\end{figure}

\begin{table*}
\footnotesize
\begin{center}
\begin{tabular}{|c|c|c|c|c|c|}
\hline
\thead{Tracker} & \thead{Avg. no. of \\ recoveries } & \thead{Avg. no. \\ of chances} & \thead{Sequences with\\ static recoveries} & \thead{Performance on\\ sequences with\\ static recoveries \\ (Success Metric)} & \thead{Reduced performance\\ on sequences with\\ static recoveries \\ (Success Metric)} \\
\hline
SPLT~\cite{iccv19_SPLT} & 0.28 & 1.2 & 5 & 29.64 & 4.83\\
\hline
MBMD~\cite{zhang2018learning} & 0.14 & 1.3 & 4 & 15.65 & 10.2 \\
\hline
ATOM~\cite{Danelljan_2019_CVPR} & 0.98 & 8.06& 12 & 25.08 & 16.12 \\
\hline
SiamRPN~\cite{Li_2018_CVPR} & 0.58  &2.22 & 9 & 40.00 & 21.22\\
\hline
MDNet~\cite{nam2016learning} & 3.16   & 15.64& 13 & 15.14 & 10.64\\
\hline
FuCoLoT~\cite{lukevzivc2018fucolot} & 2.7 & 6.78 & 17 & 10.77 & 4.29 \\
\hline
CMT~\cite{nebehay2015clustering} & 5.24 & 11.16 & 21 & 8.68 & 5.60 \\
\hline
ECO~\cite{danelljan2017eco} &  4  & 24.92  & 19& 8.52 & 5.13\\
\hline
LCT~\cite{ma2015long} & 3.34 & 7.18 &  21 & 9.66 & 7.15\\
\hline
TLD~\cite{kalal2011tracking} & 2.56  & 5.32 & 16 & 7.35 & 2.37 \\

\hline
\end{tabular}
\end{center}
\caption{Results for the analysis of static recoveries.}
\label{table:table3}
\end{table*}

The second type of recoveries we study is when the tracker is stationary, and the target passes through it, and then the tracking resumes. An example of such a recovery is illustrated in Figure~\ref{fig:static}. Here, the recovery can be attributed to chance, because the target, fortunately, moved into the tracker (the tracker recovers even though it had no idea where the target was). 

We first formalize the notion of the tracker being ``stationary." A tracker is said to be stationary if the IoU of the current prediction (at time t) is more than 0.5 with each of the previous 200 predictions and the IoU with the target is zero. This definition ensures that the tracker is frozen somewhere in the background, after accounting for minor noisy movements. We further define ``static recovery," i.e. the recovery which happens when the tracker is stationary (IoU between the tracker and target goes from zero to non-zero and remains non-zero for next 60 frames). We then compute the following:

\begin{itemize}
    \item The average number of static recoveries per sequence in the dataset. 
    \item The average number of chances i.e., number of times when the tracker was stationary, and the target came towards it leading to a non zero IoU (even for a single frame).
    \item The impact of static recoveries on the tracking performance i.e., the reduced success metric by ignoring the performance after the first static recovery in each sequence. However, here we report the performance drops only on the sequences where static recovery occurs (which differs for each tracker). The point of reporting these performance drops is not to give a metric, but to understand the worst-case impact of such recoveries on the tracking performance. 
\end{itemize}

\begin{figure*}
\begin{center}
\includegraphics[width=\linewidth]{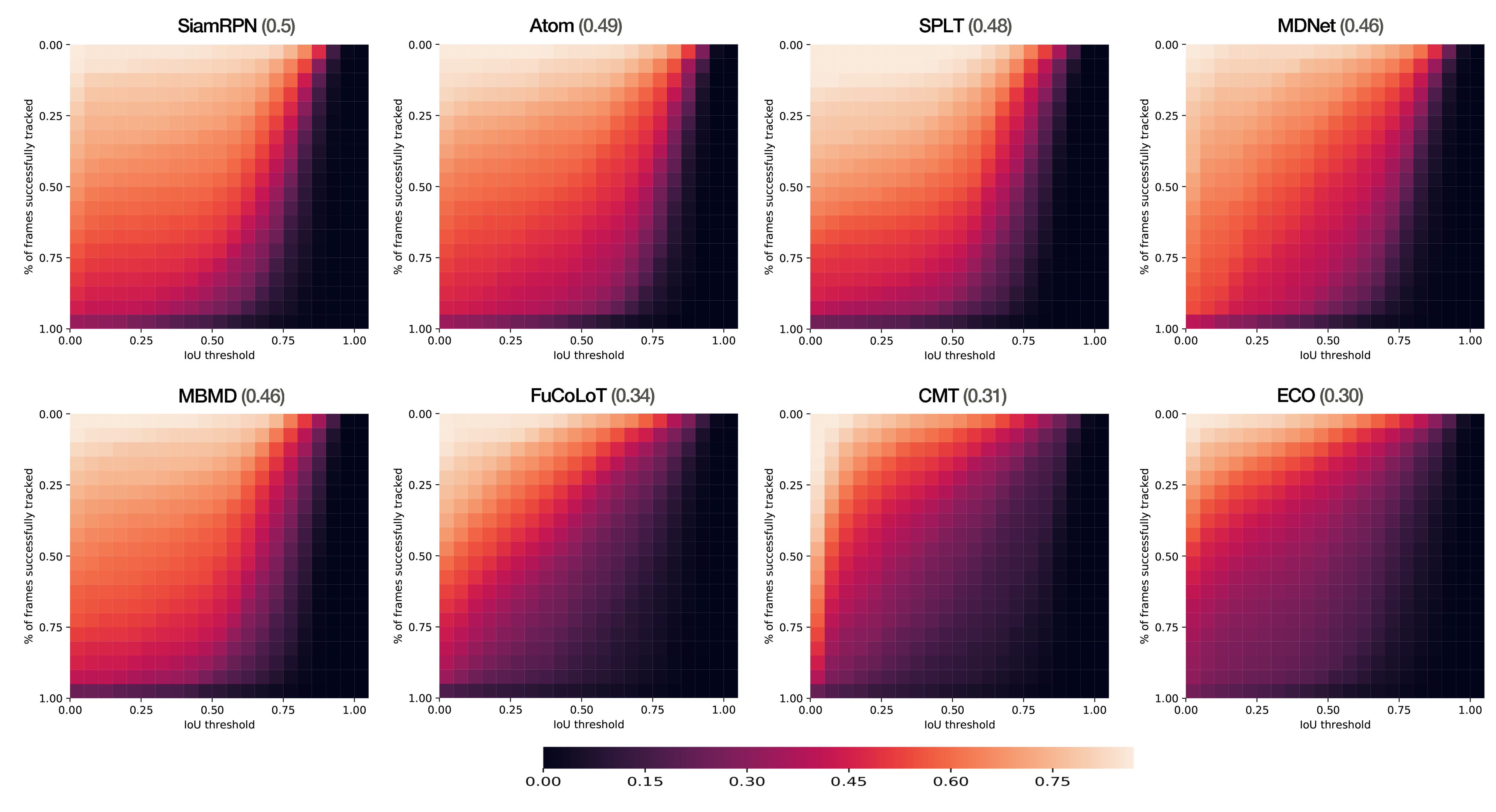}
\end{center}\vspace{-1.5em}
\caption{3D-LSM visualizations for the evaluated trackers. 3D-LSM metric is also reported for each tracker (on top).}
\label{fig:lsm}
\end{figure*}
\textbf{Results and Discussion:} The results are summarized in Table~\ref{table:table3}. The first two columns present the average number of static recoveries per sequence and the number of chances it got (averaged over all 50 sequences). The third column presents the number of sequences for each tracker which have static recoveries (the experiments are performed on all 50 sequences of the dataset; however, not all sequences have static recoveries). The last two columns present the success metric before and after accounting for the chance based recoveries (averaged only over the sequences with static recoveries, which is different for each tracker). 
Our observations are as follows:
\begin{enumerate}
    \item Trackers that perform online model updates (ECO, CMT, TLD) are prone to freezing very often. This occurs even in the case of trackers like MDNet and FuCoLoT, which only perform conservative model updates (when confident). Model updates could enable the tracker to adapt to the background and hence causing the tracker to freeze.
    \item The experiment is quantifying cases when the tracker has failed, and the tracker predictions have frozen entirely. Despite having a very strict definition that gives the benefit of the doubt to the trackers, we still observe that a lot of trackers freeze.
    \item We observe a complementary nature of recoveries. Predominantly offline trained trackers tend to look for objectness and can track an alternate object altogether. Due to the interactions between the objects, the tracker recovers. The second class of trackers which perform online model updates can sometimes lose the discriminative ability between the target and background and can freeze while tracking the background. The recovery occurs when the target passes through the tracker. 
    \item The performance drop in SPLT, MBMD, ATOM and SiamRPN is significant after accounting for the performance due to chance. This also indicates that they make good use of the chances they get. 
\end{enumerate}

\section{Reliability in Long-term Tracking}

Practically, trackers are reliable to use in long-term applications if the human effort to fix the incorrect tracker predictions is minimal. The human effort is a function of the precision required for the application at hand. A tracker which gives contiguous segments of precise tracking would be easier to correct by re-initializing on failures. However, it will take a lot of mental burden to correct a tracker whose IoU fluctuates intermittently. Moudgil and Gandhi~\cite{moudgil2017long} made an effort to quantify the reliability aspect and proposed the Longest Subsequence Measure (LSM) metric. In this section, we address some of limitations of LSM metric and extend it in a more general sense. We also present a visual interpretation of trackers which could aid the practitioner to pick appropriate trackers conditioned on their specific needs.
\begin{table}
\footnotesize
\begin{center}
\begin{tabular}{|c|c|c|c|}
\hline
\thead{Tracker} & \thead{Success Metric at IoU 0.5}\\
\hline
SPLT~\cite{iccv19_SPLT} & \textbf{52.74}\\
\hline
SiamRPN~\cite{Li_2018_CVPR} & 51.52\\ 
\hline
MBMD~\cite{zhang2018learning} & 48.12 \\
\hline
ATOM~\cite{Danelljan_2019_CVPR} & 47.51\\
\hline
MDNet~\cite{nam2016learning} & 42.27\\
\hline
FuCoLoT~\cite{lukevzivc2018fucolot} & 21.99 \\
\hline
CMT~\cite{nebehay2015clustering} & 20.81\\
\hline
ECO~\cite{danelljan2017eco} &  21.94\\
\hline
TLD~\cite{kalal2011tracking} & 13.90\\
\hline
LCT~\cite{ma2015long} & 8.75\\
\hline
\end{tabular}
\end{center}
\caption{Success Metric for the trackers on entire TLP dataset.}
\label{table:table4}
\end{table}

\textbf{Preliminaries:} LSM~\cite{moudgil2017long} computes the ratio of the length of the longest successfully tracked continuous subsequence to the total length of the sequence.  A subsequence is marked as successfully tracked,  if $x\%$  of frames within it have IoU $>$ 0.5,  where $x$ is a slack parameter. A representative LSM score per tracker is computed by fixing the slack parameter $x$ to 0.95 (tracking 95\% of the sub-sequence successfully). 

We believe that the choice of thresholds for IoU (0.5) and slack $x$ (0.95) in LSM does not provide a fair and complete perspective. For example, a tracker that has IoU slightly lesser than 0.5 would be penalized harshly due to binary IoU thresholding at 0.5. Prior work~\cite{russakovsky2015best} has also shown that human annotators cannot often distinguish between IoU scores of 0.3 and 0.5. In ~\cite{moudgil2017long}, the authors also present LSM plots by fixing IoU to 0.5 and varying the slack. However, such plots fails to give a holistic perspective on the simultaneous effect of changing both the IoU and the slack.

\textbf{Extending LSM:}
We present a 3D-LSM metric, which captures the effect of both precision (IOU) and failure tolerance in a connected manner. The 3D-LSM metrics is the mean of a matrix, computed by varying both the slack and the IoU parameters. Each entry in the matrix measures the longest contiguous sub-sequence (normalized) successfully tracked by fixing the IoU and slack parameters (for instance if the slack is 0.95 and IoU is 0.3, then we find the longest sub-sequence where 95\% of the frames are tracked with IoU greater than 0.3). Basically, each entry in the matrix is the LSM value computed at a specific slack and IoU threshold. In current experiment we vary both slack and IoU thresholds at a rate of 0.05 from 0.05 to 1, resulting in a 20$\times$20 matrix. One major benefit of the proposed metric is that it can be visualized as an image and makes way for a direct visual interpretation. It would aid non-expert practitioners to compare several trackers by visual inference.

\textbf{Results and Discussion:}
The 3D-LSM visualization results for the evaluated trackers on the TLP dataset are shown in Figure~\ref{fig:lsm}. SiamRPN, ATOM, SPLT, MDNet, and MBMD give better performance in comparison to the other five trackers. SPLT and MBMD are built upon the SiamRPN as the base network, and though they improve other aspects like re-detection, the reliability aspect reduces marginally. Another interesting observation is that while ECO outperforms SiamRPN on short term benchmarks like OTB100, it performs significantly worse in the presented long term setting. The reliability aspect of trackers like CMT is quite low, possibly due to drift in feature tracks. FuCoLoT was designed as a long term tracker; however, it performs poorly on the reliability aspect. MDNet performs well on the reliability aspect owing to its online updates. 

The 3D-LSM plots allow direct visual inferences: (a) brighter plots indicate better performance. We can observe how the images get darker when moving from SiamRPN to ECO. (b) Contours formed in more reliable trackers tend to stretch towards the bottom right corner. Compare SiamRPN and ECO, for instance; we can see that the shape of the contour inverts. (c) The practitioners need lies in the bottom right corner (i.e., low failure tolerance and high IoU), and most trackers are pitch black in the area. This highlights the significant challenges and opportunities which lie ahead in the area of visual object tracking to meet the application requirements.

\section{Summary and Conclusion}
In this paper, we touch upon the three crucial aspects of Re-detection, Recovery, and Reliability (3R's) for long term tracking. These aspects are not explicit in existing evaluation metrics, which makes it difficult to reason out the poor or effective performance of a particular tracker in the long term setting. The 3R analysis is aimed to bridge this gap and can categorically highlight the shortcomings of different tracking algorithms. It helps us reason out the overall performance of the tracker as well (Table~\ref{table:table4}). For instance, trackers like CMT and FuCoLoT are specifically designed for long term setting and have an explicit re-detection module; however, they lack reliability and end up giving a poor overall performance. 

Hence, definitions that restrict long term trackers to only the algorithms with re-detection capabilities~\cite{lukevzivc2018now} are limited and ignore the Recovery and Reliability aspects. Even trackers like MDNet (without explicit re-detection)  give a reasonable overall performance in long term context, owing to high reliability. Recently proposed SPLT tracker gives the best overall performance (Table~\ref{table:table4}); however, it only gives a marginal improvement over the base SiamRPN network. 3R analysis shows that SPLT improves on the re-detection aspect; however, compromises on reliability and also ends up tracking an alternate object often. Similar, specific insights can be drawn for other trackers as well and can aid in studying their strengths and weaknesses. Overall we believe 3R analysis paves the way for designing better tracking algorithms in the future.

{\small
\bibliographystyle{ieee}
\bibliography{egbib}
}

\end{document}